\documentclass[letterpaper, 10 pt, conference]{ieeeconf}  % Comment this line out if you need a4paper

\IEEEoverridecommandlockouts                              % This command is only needed if 
\usepackage{caption}
\usepackage{blindtext}
\usepackage{graphicx}
\usepackage{hyperref}
\usepackage{amsmath}
\usepackage{amssymb}
\usepackage{tabularx}
\usepackage{subcaption}
\usepackage{xcolor}
\usepackage[acronym]{glossaries}
\usepackage{wrapfig,lipsum,booktabs}

\glsdisablehyper
\newacronym{mse}{MSE}{mean squared error}
\newacronym{svd}{SVD}{Singular Value Decomposition}
\newacronym{dl}{DL}{Deep Learning}
\newacronym{gat}{GAT}{Graph Attention Network}
\newacronym{imu}{IMU}{Inertial Measurement Unit}
\newacronym{gnn}{GNN}{Graph Neural Network}
\newacronym{rte}{RTE}{Relative Translational Error}
\newacronym{rre}{RRE}{Relative Rotational Error}
\newacronym{gcn}{GCN}{Graph Convolution Network}
\newacronym{xai}{XAI}{eXplainable AI}
\newacronym{gxai}{GXAI}{Graph eXplainable AI}

\newacronym{mae}{MAE}{Mean Absolute Error}
\newacronym{jsd}{JSD}{Jensen-Shannon Divergence}
\newacronym{twgl}{TwGL}{Trustworthy Graph Learning}
\newacronym{gdl}{GDL}{Graph Deep Learning}
\newacronym{aad}{AAD}{Average Absolute Discrepancy}

\usepackage[hang,flushmargin]{footmisc}

\usepackage{xcolor}

\usepackage{siunitx}
\usepackage{mathtools}
\usepackage{physics}
\DeclarePairedDelimiterX\set[1]\lbrace\rbrace{\def\given{\;\delimsize\vert\;}#1}

\usepackage{multirow}

\usepackage{cuted}

\usepackage{cleveref}
\crefname{table}{Tab.}{Tabs.}
\crefname{figure}{Fig.}{Figs.}
\crefname{section}{Sec.}{Secs.}
\crefname{equation}{Eq.}{Eqs.}

\title{\LARGE \bf
Semantic Interpretation and Validation of Graph Attention-based Explanations for GNN Models
}
\author{Efimia Panagiotaki$^1$, Daniele De Martini$^2$, Lars Kunze$^1$\\% <-this % stops a space
$^1$Cognitive Robotics Group and $^2$Mobile Robotics Group, Oxford Robotics Institute, \\ Department of Engineering Science, University of Oxford, UK\\
\texttt{\{efimia,daniele,lars\}@robots.ox.ac.uk}
\thanks{
This work was supported by a Google DeepMind Engineering Science Scholarship, the EPSRC project RAILS (grant reference: EP/W011344/1), and the Oxford Robotics Institute research project RobotCycle.
}
}

\begin{document}

\maketitle
\thispagestyle{empty}
\pagestyle{empty}

%%%%%%%%%%%%%%%%%%%%%%%%%%%%%%%%%%%%%%%%%%%%%%%%%%%%%%%%%%%%%%%%%%%%%%%%%%%%%%%%
\begin{abstract}
In this work, we propose a methodology for investigating the use of semantic attention to enhance the explainability of \gls{gnn}-based models. \Gls{gdl} has emerged as a promising field for tasks like scene interpretation, leveraging flexible graph structures to concisely describe complex features and relationships. As traditional explainability methods used in \gls{xai} cannot be directly applied to such structures, graph-specific approaches are introduced. Attention has been previously employed to estimate the importance of input features in \gls{gdl}, however, the fidelity of this method in generating accurate and consistent explanations has been questioned. To evaluate the validity of using attention weights as feature importance indicators, we introduce semantically-informed perturbations and correlate predicted attention weights with the accuracy of the model. Our work extends existing attention-based graph explainability methods by analysing the divergence in the attention distributions in relation to semantically sorted feature sets and the behaviour of a \gls{gnn} model, efficiently estimating feature importance. We apply our methodology on a lidar pointcloud estimation model successfully identifying key semantic classes that contribute to enhanced performance, effectively generating reliable post-hoc semantic explanations.

\end{abstract}

\glsresetall

%%%%%%%%%%%%%%%%%%%%%%%%%%%%%%%%%%%%%%%%%%%%%%%%%%%%%%%%%%%%%%%%%%%%%%%%%%%%%%%%
\section{Introduction}\label{sec:introduction}
\Gls{twgl} identifies reliability, explainability, accountability, and other trust-oriented features as key requirements for trustworthy \gls{gdl} \cite{Wu2022,Zhang2022}.
Undeniably, trust is a critical design factor for the successful development and deployment of autonomous vehicles.
Trust and explainability are inherently linked. Explaining the decisions of autonomous vehicles enables users and regulatory bodies to use and work on transparent and accountable systems. Further, having a clear understanding of the capabilities and limitations of an autonomous system increases trust in the underlying technology and fosters its adoption.
% mental model of the system, and regulatory bodies to work on a transparent and accountable system. Having a clear 

 In real-world deployment, autonomous vehicles are required to navigate safely in unknown and dynamic environments.
 To ensure safe operation, the system must effectively assess the complexity of traffic scenes and make logical decisions based on its anticipated performance. 
 A critical prior requirement for reliable decision-making is for those vehicles to know their precise location relative to their observed surroundings. 
 This relates to the task of \emph{pose estimation}, which calculates the position of the ego-vehicle w.r.t. the perceived environmental features.
 
 Our proposed research focuses on analysing and explaining the complexity of the environment using learned attention weights to identify the contribution of each semantic element, i.e. static and dynamic agents and morphological structures, to the performance of a baseline lidar pointcloud-based pose estimation model.
 Similar to \cite{Fan2021}, we take inspiration from perturbation-based \gls{gxai} methods to investigate the validity of using attention weights as feature-importance indicators.
In our work, we extract semantic sets and rank them based on their attention scores. We conclude on their importance in the pose estimation task by verifying the correlation between attention weights and model accuracy. We semantically perturb the input and, as proposed in \cite{Wiegreffe2019,jain2019attention,Serrano2020}, we measure the distribution divergence to calculate the contribution of each set’s attention weights
to the overall attention distribution.
 
Our key contributions are as follows:
\begin{itemize}
    \item A methodology for assigning importance scores to semantic sets based on their contribution to the performance of a \gls{gnn} model;
    \item A semantic interpretation of learned attention weights in correlation with the predictions of a graph-based model;
    \item A semantically-informed perturbation process for evaluating the explanations for \gls{gxai}.
    %\item
\end{itemize}
The model used as our baseline is a graph-attention-based pose estimation model, SEM-GAT \cite{panagiotaki2023semgat}, trained on the KITTI Odometry Dataset \cite{6248074}. Our method examines the interpretability of the model w.r.t. its output predictions, eventually assessing its efficacy in real-world applications.
%%%%%%%%%%%%%%%%%%%%%%%%%%%%%%%%%%%%%%%%%%%%%%%%%%%%%%%%%%%%%%%%%%%%%%%%%%%%%%%%
\begin{figure*}[t]
\vspace{1.5mm}
\centering
\includegraphics[width=0.8\textwidth]{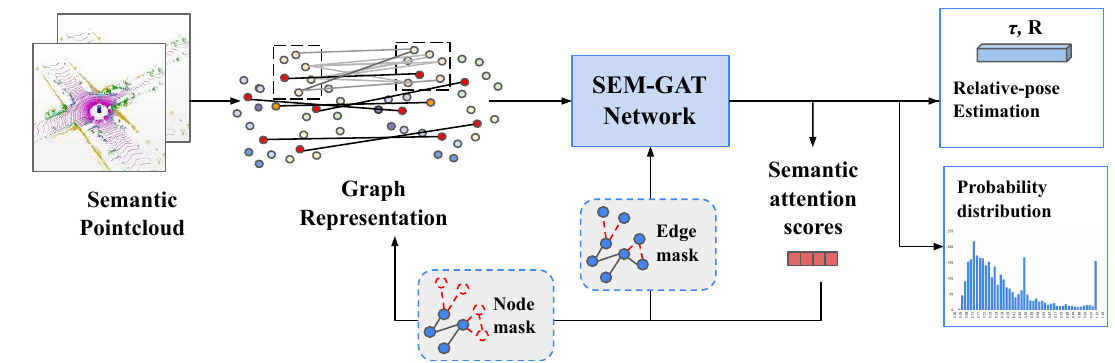}
\caption{Overview of our proposed methodology. After retrieving the attention weights for each semantic class from vanilla SEM-GAT, we use a node mask to perturb the model's input by masking the highest-ranking semantic class set to measure the divergence in the distribution of the attention weights. We correlate this measurement with the pose estimation error from masked SEM-GAT to estimate importance scores for each semantic set. We repeat this process by masking the last layer of the model using an \emph{edge mask}.
% an \emph{edge mask} to mask the highest ranking semantic class sets at the last layer of the model and measure the divergence of the attention weights distribution. We correlate this measurement with the pose estimation error from masked SEM-GAT to generate importance scores for each semantic set. We repeat this process, perturbing the model's input graphs using a \emph{node mask}.
}
\label{fig:overview}
\vspace{-0.5cm}
\end{figure*}

\section{Related Work}\label{sec:related_work}
Recent studies have investigated the topic of explainability in \glspl{gnn} proposing different approaches to explain their predictions.
Following the taxonomy for instance-level explanations introduced in \cite{Yuan2022}, these methods can be categorised into gradient/feature-, decomposition-, surrogate-, and perturbation-based. 

Gradient/feature-based methods \cite{SA_GuidedBP,GradCAM} calculate the gradients of the output with respect to the extracted features in the input via backpropagation and use them to estimate attribution scores.
Decomposition-based methods \cite{GradCAM,lrp,GNN-LRP} estimate the importance scores by expanding the network inference blocks into a sum of effects and identifying structures that contribute to the prediction.
Surrogate-based methods \cite{graphlime,relex,vu2020pgmexplainer} use simpler, interpretable models to approximate explanations on the original ones.
Perturbation-based methods \cite{GNNExplainer,pgexplainer,zorro,graphmask,casual,SubGraphX} measure importance scores by iteratively masking the input and calculating the changes in the output predictions, generating post-hoc explanations. Perturbation-based methods are the most relevant to our approach.
Whereas these methods rely on random masks to perturb the input, we argue for more effective and concise perturbations by conditioning the masks on the input. Specifically, our proposed methodology generates semantics-driven masks to perturb the input of a \gls{gnn} model and generate explanations for its output predictions.
% In our proposed research, the predicted attention weights correspond to the estimated importance scores of the input features, the accuracy of which is then evaluated using semantics to inform the masking on input perturbations. Our work can easily be extended to provide semantic explanations at the inference step.

Attention has been employed to interpret how input features influence the predictions of deep learning models \cite{spatiotemporal_rnn, Zuo2022, SCARLET}.
However, these holistic explanations have previously been regarded as insufficient and inaccurate \cite{jain2019attention, exploring_attention}.
Later studies \cite{Wiegreffe2019,Serrano2020} challenged this stark position by reclaiming attention's role as an explainability method, albeit with limitations in terms of accuracy and applicability. Combining attention weights with the models' properties has been shown to produce more reliable and consistent explanations \cite{exploring_attention}.
Building upon this rationale, \cite{Fan2021} suggests that a verification step can prove the relation between attention weights and feature importance by correlating the effects of different attention weight distributions to the accuracy of the models. This can be a prerequisite step before employing attention to explain the model's performance.

Following these studies, we evaluate the validity of using graph attention to generate explanations by correlating the accuracy of a baseline model with the distribution divergence of attention weights after iterative perturbations.
Our results demonstrate that attention can be useful to identify important semantics in the environment that contribute towards reliable model performance.
%for our model attention can be useful for explainability. 

\section{Preliminaries}\label{sec:overview}

In this section, we formulate the problem addressed and describe the graphs and the \gls{gnn} model used as the baseline.

\subsection{Problem Definition and Notations}\label{subsec:problemdef}
Let $P_{t} : \set{\mathbf{p}_i \given \mathbf{p}_i \in \mathbb{R}^3}$  be a pointcloud at discrete timestamp $t$ in a total of $N$ consecutive scans.
$P_t$ can be subdivided into a set of semantic classes $\mathbb{S}$ that may include \emph{terrain}, \emph{buildings}, \emph{trees}, \emph{vehicles}, and \emph{pedestrians}, among others. For each point $\mathbf{p}_i$, we assign a semantic label $s_i \in \mathbb{S}$.
In our proposed work, we aim to identify the most significant semantic classes for accurate relative pose estimation between two consecutive pointclouds, $P_t$ and $P_{t+1}$. Here, we denote the relative pose as $[\mathbf{R}_{t,t+1} | \mathbf{\boldsymbol\tau}_{t,t+1}]$, where $\mathbf{R}_{t,t+1} \in \mathbb{SO}(3)$ is the rotation and $\mathbf{\boldsymbol\tau}_{t,t+1} \in \mathbb{R}^3$ is the translation.

\subsection{SEM-GAT}\label{subsec:sem_gat}
\begin{figure}
\vspace{1.5mm}
\centering
\includegraphics[width=\columnwidth]{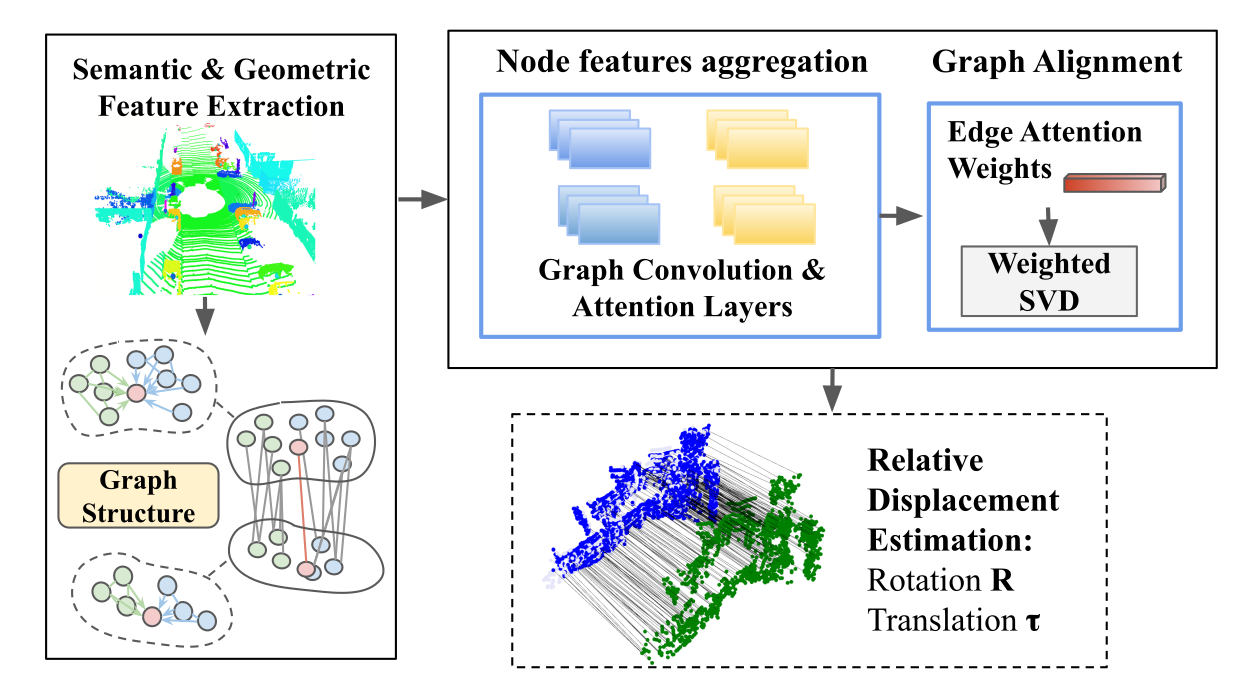}
\caption{SEM-GAT, the attention-based \gls{gnn} used as baseline for generating and validating semantic explanations.
}
\label{fig:semgat}
\vspace{-0.5cm}
\end{figure}

We employ SEM-GAT \cite{panagiotaki2023semgat} for generating attention-based explanations.
SEM-GAT is a semantic graph-based pose estimation \gls{gnn} model, depicted in \cref{fig:semgat}.
It estimates the relative transformation between two pointclouds by identifying potential matching correspondences between those pointclouds for registration. SEM-GAT explicitly employs attention to weigh each candidate matching pair, making it a suitable baseline to test our evaluation methodology.

SEM-GAT's input is a static graph structure $G_t$ comprising the two pointclouds $P_t$ and $P_{t+1}$.
We define the input graphs as $G_t = \langle V_t, E_t \rangle$, where $V_t$ and $E_t$ are the sets of nodes and edges, respectively.
% Given $P_{t}$ and $P_{t+1}$, we construct a static graph representation $G_{k}$ of the two pointclouds by semantically linking the nodes in the graph to generate a graph-structure representation of the environment.
Each point $\mathbf{p}_i \in P_t$ and $\mathbf{p}_j \in P_{t+1}$ is a node in $V_t$ and the edges correspond to the semantic relationships between the points according to their associated semantic label $s_i \in \mathbb{S}$ and their geometric characterisation as \emph{corner} or \emph{surface} points.
For the sake of the notation, from now on, we will drop the subscript $t$ for the instant in time.

Notably, $C \subset E$ is the set of registration-candidate pairs -- where $c_{ij} \in C$ links $\mathbf{p}_i \in P_t$ and $\mathbf{p}_j \in P_{t+1}$ -- which SEM-GAT uses to estimate the relative-pose transformations $[\hat{\mathbf{R}}_{t, t+1} | \hat{\mathbf{\boldsymbol\tau}}_{t, t+1}]$. % by finding strong registration candidates in the embedding representations of the nodes of the input graph.
The model generates feature embedding representations, encoding structural and semantic information through convolutions and multi-head graph attention \cite{velivckovic2017graph}. It then assigns attention weights $\alpha_{ij} \in \mathbb{R}$ as confidence scores to edges connecting potential registration candidate pairs $c_{ij}$.
The scores $A : \set{ \alpha_{ij} }$ are used as weights in a \gls{svd} module to align the pointclouds and recover their relative transformation.

\section{Attention-based Semantic Explanations}\label{sec:explanations}
\Cref{fig:overview} depicts the overview of our pipeline. We estimate the importance of the semantic elements in the environment using the attention weights $A$ predicted in the last layer of SEM-GAT.
To validate the suitability of using attention to explain the performance of SEM-GAT semantically, we iteratively perturb the input, correlating the attention-weights-distribution divergence with the changes in the model's accuracy.

\begin{figure}
\vspace{1.5mm}
\centering
\begin{subfigure}{0.44\textwidth}
  \includegraphics[width=\columnwidth]{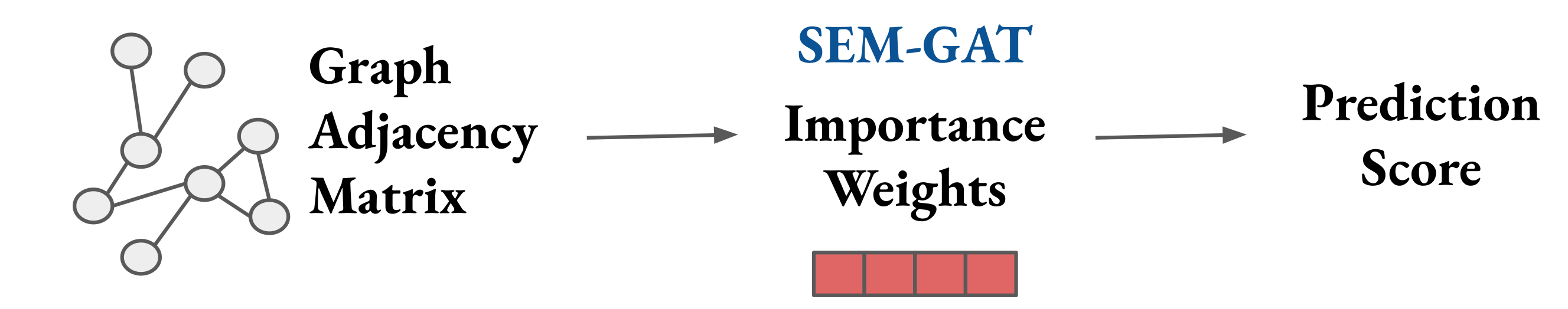}
  \caption{Average attention scores as importance weights.}
  % \vspace{3.5mm}
\end{subfigure}

\begin{subfigure}{0.46\textwidth}
  \includegraphics[width=\columnwidth]{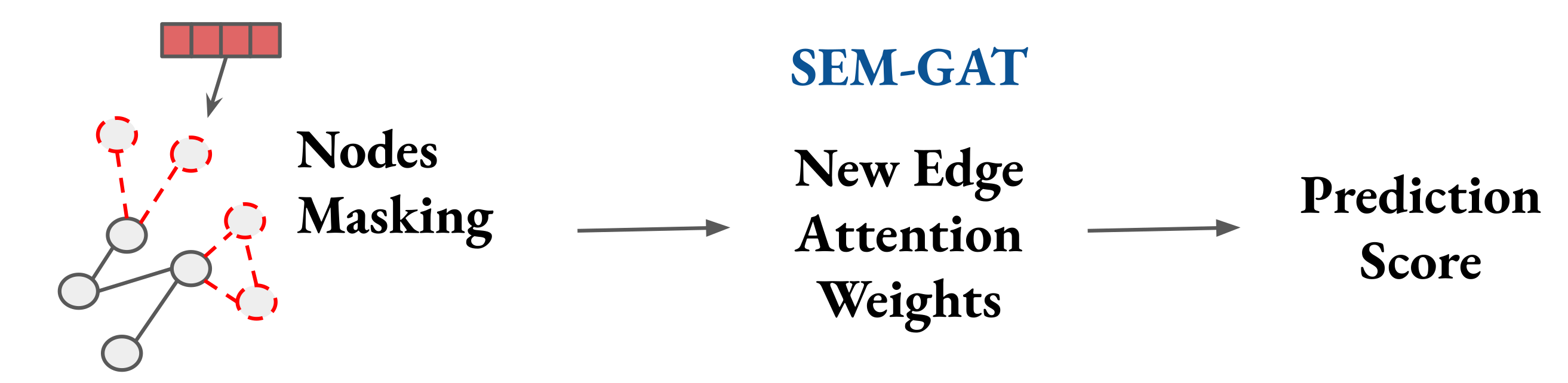}
  \caption{Node masking in the input graphs.}
  \vspace{1mm}
\end{subfigure}

\begin{subfigure}{0.46\textwidth}
  \includegraphics[width=\columnwidth]{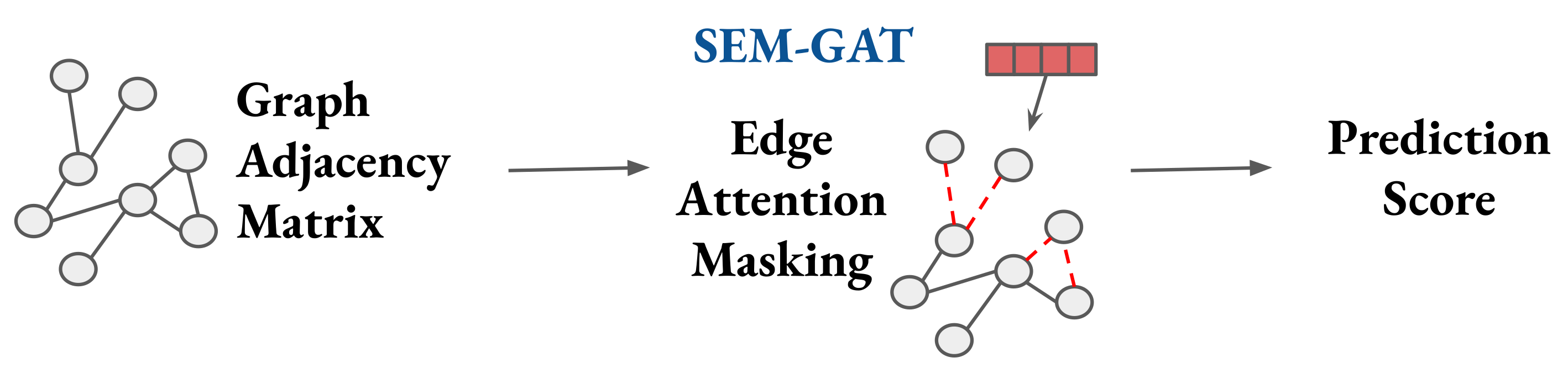}
  \caption{Edge masking at the last layer of SEM-GAT.}

\end{subfigure}
% \caption{Overview of the perturbation process as Input $\rightarrow$ Model $\rightarrow$ Output.}
 \caption{Overview of the perturbation process as \emph{Input} $\rightarrow$ \emph{Model} $\rightarrow$ {Output}: (a) visualises the process of extracting the semantic importance weights from vanilla SEM-GAT. These weights then inform the two independent steps of (b) node- and (c) edge-attention-weights masking.}
\label{fig:pert}
    \vspace{-3mm}

\end{figure}

We first investigate the semantic interpretation of the attention weights $A$ by ranking the semantic classes at inference according to their predicted total weights, normalised on the number of points.
Based on this ranking, we extract semantic feature sets to iteratively mask the model while measuring the output variations.
% We investigate the semantic interpretation of the attention scores from SEM-GAT. In particular, due to the fact that edge attention weights are calculated based on the node embeddings derived from nearest neighbor aggregations, as described in \cref{subsec:sem_gat}, we investigate whether the morphological features of each class generate more concise and descriptive embeddings which result in them receiving highest importance scores, and thus making the query class more important than others.
We propose two different methodologies, visualised in \cref{fig:pert}:
\begin{enumerate}
    \item Masking the nodes of the input graph according to the average overall attention score of the semantic sets calculated in post-processing. This effectively alters the elements \textit{and} context of the input.
    \item Zeroing the edge attention weights of our estimated most important semantic sets at the last layer of SEM-GAT, directly masking the edges with the highest confidence weights for SVD.
\end{enumerate}

Following the outcome of the perturbations, we evaluate the adequacy of using attention weights as importance indicators.
The validation process can be split into two parts: 1) measuring the attention distribution divergence and 2) correlating the attention scores with the model's performance before and after masking.

\subsection{Attention Distribution Divergence}\label{subsec:distributions}
Given a graph $G$, we follow the works \cite{Wiegreffe2019, jain2019attention, Serrano2020} and calculate the variation in the predicted attention scores caused by the perturbation with the \gls{jsd} distance.
Defining $\alpha^b$ and $\alpha^a$ as the distributions of weights before and after the perturbation, with $b$ corresponding to vanilla SEM-GAT:
\begin{equation}\label{eq:jsd}
    JSD(\alpha^{b}, \alpha^{a}) = \sqrt{{\frac{D_{KL}(\alpha^{b} \parallel \bar{\alpha}) + D_{KL}(\alpha^{a} \parallel \bar{\alpha})}{2}}}
\end{equation}
$0 \leq JSD(\cdot, \cdot) \leq 1$. $D_{KL}$ corresponds to the Kullback-Leibler divergence and $\bar{\alpha}$ is the distribution of attention weights averaged edgewise between before and after perturbation.
We consider the total \gls{jsd} of a sequence of pointclouds as the average of the \glspl{jsd} on the sequence.

% $D_{KL}$ corresponds to the Kullback-Leibler divergence and $\bar{\alpha}$ is the pointwise mean of $\alpha_{k_b}$ and $\alpha_{k_a}$. The JSD distance corresponds to the square root of the \gls{jsd} metric, used in \cite{Wiegreffe2019, jain2019attention, Serrano2020}.

\subsection{Attention-Performance Correlation}\label{subsec:correlation}
As we mask the graph, we measure the variations in SEM-GAT's pose estimation accuracy to assess the correlation between attention and model performance.
The authors in \cite{Fan2021} propose using the discrepancy in the model's accuracy before and after masking; similarly, we calculate the \gls{aad} of an accuracy score $\hat{\mathbf{y}}$ from before and after masking as:
\begin{equation}\label{eq:e}
    AAD(\hat{\mathbf{y}}^{b}, \hat{\mathbf{y}}^{a}) = |\hat{\mathbf{y}}^b - \hat{\mathbf{y}}^a|
\end{equation}
This metric is a good indicator of the fluctuations in the output predictions in each perturbation step. 

For our case, we consider as $\hat{\mathbf{y}}$ the \gls{rre} [\si{\degree}] and \gls{rte} [\si{\metre}] between SEM-GAT's rotation and translation estimations $[\hat{\mathbf{R}}|\hat{\mathbf{\boldsymbol\tau}}]$ and ground-truth values $[\mathbf{R}|\mathbf{\boldsymbol\tau}]$, respectively.
The two metrics are defined as:
\begin{equation}\label{eq:rre}
RRE = \acos(\frac{1}{2} (\mathrm{tr}(\mathbf{R}^\top \hat{\mathbf{R}}) - 1))
\end{equation}
\begin{equation}\label{eq:rte}
RTE = \norm{\mathbf{\boldsymbol\tau}_{gt} - \hat{\mathbf{\boldsymbol\tau}}}_2
\end{equation}

We consider the total \gls{aad} over an entire sequence as the average of the two metrics' \glspl{aad}.
The combined average absolute discrepancy \gls{aad} is then calculated as follows:
\begin{multline}\label{eq:e_total}
    AAD = \frac{\sum_{t = 1}^{N-1}|RRE_{b} - RRE_{a}|}{N-1} + \\
    \frac{\sum_{t = 1}^{N-1}|RTE_{b} - RTE_{a}|}{N-1}
\end{multline}

\section{Results}\label{sec:results}

SEM-GAT is trained and evaluated on Sequences \texttt{00}, \texttt{02}, and \texttt{03} of the KITTI Odometry Dataset \cite{6248074}. We test our approach on every sequence in the dataset, from \texttt{00} to \texttt{10}.
We use the ground-truth labels and poses from SemanticKITTI \cite{Behley2019} to generate our semantic graphs and evaluate the performance of SEM-GAT by correlating \gls{aad} with \gls{jsd} to estimate the contribution of the query semantic importance scores to the accuracy of the model.

\subsection{Semantic Masking}\label{subsec:masking}
We use the predicted attention weights from the last layer of SEM-GAT to rank the semantic classes in the dataset according to their average learned attention scores for each sequence.
\Cref{table:semantic_order} reports each sequence's five most important semantic classes and their average attention values.
% \newcolumntype{?}{!{\vrule width 2pt}}

\begin{table*}
\vspace{1.8mm}
    \centering
    \renewcommand{\arraystretch}{1.2}
    \begin{tabular}{|r||lll|}
    \hline
    \multicolumn{1}{|l||}{\textbf{}} & \multicolumn{3}{l|}{\textbf{Random Classes; Average Attention Scores in Descending Order($\rightarrow$)}}                          \\ \hline
    \texttt{00}                      & \multicolumn{1}{l}{vegetation (0.38)}   & \multicolumn{1}{l}{trunk (0.34)}         & terrain (0.29)      \\ \hline
    \texttt{01}                      & \multicolumn{1}{l}{terrain (0.39)}      & \multicolumn{1}{l}{car (0.29)}           & other-ground (0.18) \\ \hline
    \texttt{02}                      & \multicolumn{1}{l}{car (0.34)}          & \multicolumn{1}{l}{building (0.33)}      & terrain (0.31)      \\ \hline
    \texttt{03}                      & \multicolumn{1}{l}{car (0.3)}           & \multicolumn{1}{l}{building (0.28)}      & traffic-sign (0.15) \\ \hline
    \texttt{04}                      & \multicolumn{1}{l}{trunk (0.33)}        & \multicolumn{1}{l}{building (0.2)}       & traffic-sign (0.18) \\ \hline
    \texttt{05}                      & \multicolumn{1}{l}{trunk (0.31)}        & \multicolumn{1}{l}{pole (0.29)}          & other-vehicle (0.1) \\ \hline
    \texttt{06}                      & \multicolumn{1}{l}{other-ground (0.27)} & \multicolumn{1}{l}{truck (0.16)}         & bicycle (0.14)      \\ \hline
    \texttt{07}                      & \multicolumn{1}{l}{bicycle (0.11)}      & \multicolumn{1}{l}{other-vehicle (0.1)}  & traffic-sign (0.1)  \\ \hline
    \texttt{08}                      & \multicolumn{1}{l}{person (0.17)}       & \multicolumn{1}{l}{other-vehicle (0.16)} & traffic-sign (0.12) \\ \hline
    \texttt{09}                      & \multicolumn{1}{l}{car (0.34)}          & \multicolumn{1}{l}{traffic-sign (0.12)}  & person (0.1)        \\ \hline
    \texttt{10}                     & \multicolumn{1}{l}{trunk (0.25)}        & \multicolumn{1}{l}{other-ground (0.04)}  & person (0.1)        \\ \hline
            \multicolumn{4}{c}{\textbf{   }}

    \end{tabular}    
    
    \begin{tabular}{|l||lllll|}
    \hline
        ~ & \multicolumn{5}{c|}{\textbf{Highest Ranking Classes; Average Attention Scores in Descending Order ($\rightarrow$)}}  \\ \hline
        \texttt{00} &pole $(0.55)$&sidewalk $(0.53)$&fence $(0.44)$&building $(0.4)$&bicycle $(0.4)$\\ \hline
        \texttt{01} &fence $(0.51)$&vegetation $(0.42)$&terrain $(0.39)$&car $(0.29)$&ground $(0.18)$\\ \hline
        \texttt{02} & sidewalk $(0.56)$&fence $(0.48)$& trunk $(0.45)$& vegetation $(0.4)$& pole $(0.36)$\\ \hline
        \texttt{03} & pole $(0.55)$&sidewalk $(0.55)$&fence $(0.5)$&vegetation $(0.38)$&terrain $(0.38)$\\ \hline
        \texttt{04} &sidewalk $(0.6)$&pole $(0.49)$&fence $(0.45)$&car $(0.44)$&vegetation $(0.43)$\\ \hline
        \texttt{05} &sidewalk $(0.56)$&terrain $(0.5)$&fence $(0.47)$&car $(0.4)$&building $(0.4)$\\ \hline
        \texttt{06} &pole $(0.6)$&sidewalk $(0.57)$&trunk $(0.52)$&terrain $(0.45)$&car $(0.45)$\\ \hline
        \texttt{07} &pole $(0.56)$&sidewalk $(0.54)$&fence $(0.46)$&building $(0.4)$&car $(0.39)$\\ \hline
        \texttt{08} &sidewalk $(0.55)$&pole $(0.51)$&terrain $(0.43)$&trunk $(0.42)$&building $(0.4)$\\ \hline
        \texttt{09} &sidewalk $(0.55)$&terrain $(0.44)$&trunk $(0.43)$&vegetation $(0.39)$&fence $(0.38)$\\ \hline
        \texttt{10} &pole $(0.49)$&fence $(0.47)$&sidewalk $(0.44)$&vegetation $(0.38)$&building $(0.37)$\\ \hline 

    \end{tabular}

    \caption{Attention-based importance ranking of semantic classes in Sequences \texttt{00} to \texttt{10} of SemanticKITTI \cite{Behley2019}. This ranking guides the perturbations. Seq. \texttt{00}, \texttt{02}, and \texttt{06} to \texttt{09} were captured in urban environments, Seq. \texttt{03} to \texttt{05} and \texttt{10} in the countryside, and Seq. \texttt{01} in a highway.}
    \label{table:semantic_order}
    \vspace{-0.5cm}
\end{table*}

According to the ranking in \cref{table:semantic_order}, we split and perturb the input data in the following semantic sets:
\begin{itemize}
    \item Single-class; separately masking the top $3$ highest-scoring classes.
    \item Multi-class; masking the top $3$ and top $5$ highest-scoring classes, as well as $3$ random classes.
    \item Single-feature; masking corner or surface points.
\end{itemize}
We then evaluate whether the attention weights of these sets represent key semantic structures in the environment based on their contribution to SEM-GAT's performance.

\subsection{Attention-JSD Correlation}\label{subsec:res_ad}
\begin{figure}[h]
    \centering
    \includegraphics[width=\columnwidth, keepaspectratio]{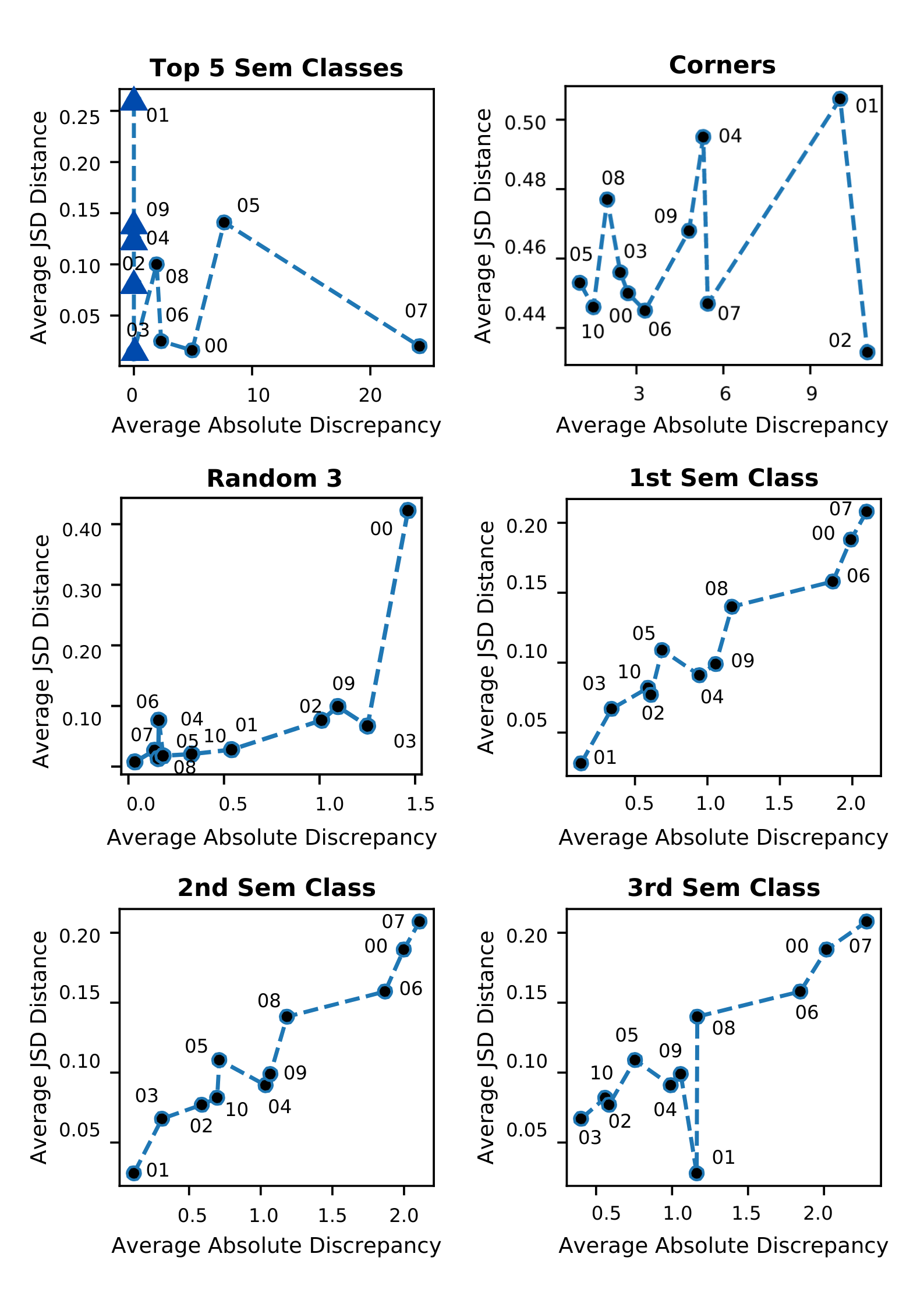}
    \vspace{-0.7cm}
    \caption{Average \gls{jsd} distance correlation with the average absolute discrepancy \gls{aad}, calculated after perturbing the last layer of SEM-GAT for Seq. \texttt{00} to \texttt{10} in SemanticKITTI.}
\vspace{-0.5cm}
\label{fig:jsd_errors}
\end{figure}

To estimate the contribution of each masking set to the total distribution of attention weights predicted in the last layer of SEM-GAT, we calculate the \gls{jsd} distance of the distributions before and after removing the weights corresponding to each set during node masking.
Higher \gls{jsd} values correspond to a larger overall contribution of the query set of semantic attention weights to the total distribution of attention. As can be seen in the vertical axis in \cref{fig:jsd_errors}, the attention weight masking sets \texttt{corner}, \texttt{5-class}, and \texttt{Random 3-class} produce higher overall \gls{jsd} scores compared to the scores from \texttt{single-class} masking. Similarly, the attention weights in the \texttt{corner} set have the highest probability density corresponding to almost half the distribution. 

\subsection{JSD-Performance Correlation}
To investigate the correlation between attention weights and model performance, we retrieve the total \gls{aad} from each sequence and correlate it with the \gls{jsd} results. When masking the larger semantic sets, \texttt{corner}, \texttt{5-class}, and \texttt{Random 3-class}, AAD fluctuations are also caused by the large number of points masked. Thus, we are mainly interested in the \texttt{single-class} masking sets in which the number of points masked is negligible compared to the entire pointcloud. Consequently, any AAD fluctuation is caused by the divergence in the attention weights' distribution and not the downsampling of the pointcloud.

As can be seen in \cref{fig:jsd_errors}, there is a strong linear correlation between \gls{jsd} and \gls{aad} for all \texttt{single-class} sets. Our results indicate that \gls{aad} is proportional with \gls{jsd} on every masking set, proving the validity of using attention weights as importance indicators and correlating them with the changes in the model's performance.

To investigate this further, for each sequence, we compare the results in \cref{table:TVE_Adj} and correlate them with \cref{fig:jsd_errors}. For all sequences, except Seq. 4, the ranking of AAD scores is proportional to the ranking in \gls{jsd} scores. For example, in Seq. 00 we observe the highest AAD and highest \gls{jsd} when masking the first semantic class. These results indicate that the semantic sets with the highest \gls{jsd} are the most important for SEM-GAT. In the \texttt{single-class} sets, there is no clear ordering relationship between them. This is expected because, as seen in \cref{table:semantic_order}, their average attention scores are very similar. 

\subsection{Qualitative Discussion}
% Following the preceding analysis, SEM-GAT is expected to perform best in semantically-rich environments with great morphological diversity.
Notably, the SemanticKITTI dataset \cite{Behley2019} is particularly interesting due to its diverse domain coverage.
It is then well-suited to analyse how SEM-GAT performs in different environments.
On every urban or countryside sequence, the highest ranking class is either \texttt{pole} or \texttt{sidewalk} as seen in \cref{table:semantic_order}. In Seq. 01, captured in a highway, such semantics are not observed and thus, naturally, the class \texttt{fence} is found to be the most important one. Moreover, we observe higher attention scores assigned to corner points than to surface points, further justifying the high ranking of classes like \texttt{vegetation, sidewalk,} and \texttt{fence} which mainly consist of corner points. 

It is particularly interesting that large semantic classes like \texttt{building} are not assigned high-importance scores from the model. After examining the model's architecture, we conclude that SEM-GAT assigns lower overall attention scores to these semantic classes due to the dense distribution of points in each instance, making it challenging for the model to identify the most important segments within them.

\begin{table*}[h]
\vspace{1.8mm}
\centering
 \resizebox{0.8\textwidth}{!}{    
 \begin{tabular}{c|c|c|c|c|c|c|c|c}
    %\hline
        \multirow{2}{*}{\textbf{seq}} & \multicolumn{8}{c}{\textbf{Average Absolute Discrepancy: Nodes Masking}} \\ 
        \textbf{ } & \textbf{Top 5 Classes} & \textbf{Top 3 Classes} & \textbf{Random 3 Classes} & \textbf{Surfaces} & \textbf{Corners} & \textbf{1st Class} & \textbf{2nd Class} & \textbf{3rd Class} \\\hline
        00 & \color{blue}\textbf{4.932} & 1.019 & 1.463 & 3.807 & \color{red}\textbf{5.214} & 1.99 & 1.998 & \color{violet}\textbf{2.017} \\ 
        01 & \color{blue}\textbf{---} & \color{blue}\textbf{---} & 0.540 & 8.688 & \color{red}\textbf{9.162} & 0.128 & 0.115 & \color{violet}\textbf{0.162} \\ 
        02 & \color{blue}\textbf{---} & \color{blue}\textbf{---} & 1.011 & \color{red}\textbf{9.316} & 2.043 & \color{violet}\textbf{0.61} & 0.589 & 0.585 \\ 
        03 & \color{blue}\textbf{---} & 1.888 & 1.251 & 3.629 & \color{red}\textbf{9.378} & 0.339 & 0.311 & \color{violet}\textbf{0.401} \\ 
        04 & \color{blue}\textbf{---} & 2.424 & 0.181 & 5.538 & \color{red}\textbf{7.228} & 0.945 & \color{violet}\textbf{1.034} & 0.991 \\ 
        05 & \color{blue}\textbf{7.64} & 2.592 & 0.137 & 2.698 & \color{red}\textbf{6.970} & 0.687 & 0.711 & \color{violet}\textbf{0.756} \\ 
        06 & \color{blue}\textbf{2.308} & 0.903 & 0.159 & \color{red}\textbf{4.196} & 3.734 & 1.865 & \color{violet}\textbf{1.867} & 1.844 \\ 
        07 & \color{blue}\textbf{24.263} & 0.837 & 0.034 & 5.642 & \color{red}\textbf{6.305} & 2.099 & 2.108 & \color{violet}\textbf{2.281} \\ 
        08 & \color{blue}\textbf{1.929} & 0.374 & 0.156 & 3.329 & \color{red}\textbf{4.162} & 1.169 & \color{violet}\textbf{1.183} & 1.166 \\ 
        09 & \color{blue}\textbf{---} & 2.22 &  1.096 & \color{red}\textbf{5.211} & 1.506 & 1.057 & \color{violet}\textbf{1.067} & 1.057 \\ 
        10 & \color{blue}\textbf{---} & 2.314 & 0.331 & 3.010 & \color{red}\textbf{10.656} & 0.59 & \color{violet}\textbf{0.696} & 0.56 \\ %\hline
        \multicolumn{8}{c}{\textbf{   }}
    \end{tabular}}
\resizebox{0.8\textwidth}{!}{ 
\begin{tabular}{c|c|c|c|c|c|c|c|c}
    %\hline
        \multirow{2}{*}{\textbf{seq}} & \multicolumn{8}{c}{\textbf{Average Absolute Discrepancy: Edges Masking}} \\ 
        \textbf{ } & \textbf{Top 5 Classes} & \textbf{Top 3 Classes} & \textbf{Random 3 Classes} & \textbf{Surfaces} & \textbf{Corners} & \textbf{1st Class} & \textbf{2nd Class} & \textbf{3rd Class} \\ \hline
         00 & \color{blue}\textbf{4.954} & 0.989 & 4.653 & 3.969 & \color{red}\textbf{4.653} & 1.98 & 1.967 & \color{violet}\textbf{1.99} \\ 
        01 & \color{blue}\textbf{---} & \color{blue}\textbf{---} & 8.198 &\color{red}\textbf{9.481} & 8.198 & \color{violet}\textbf{0.149} & 0.113 & 0.143 \\ 
        02 & \color{blue}\textbf{---} & \color{blue}\textbf{---} & 1.907 & \color{red}\textbf{9.826} & 1.907 & 0.593 & 0.591 & \color{violet}\textbf{0.605} \\ 
        03 & \color{blue}\textbf{---} & 1.861 & 8.047 & 4.211 & \color{red}\textbf{8.047} & 0.351 & \color{violet}\textbf{0.364} & 0.331 \\ 
        04 & \color{blue}\textbf{---} & 2.502 & 5.743 & \color{red}\textbf{5.884} & 5.743 & \color{violet}\textbf{0.948} & 0.925 & 0.855 \\ 
        05 & \color{blue}\textbf{7.605} & 2.556 & 5.935 & 2.916 & \color{red}\textbf{5.935} & 0.684 & 0.695 & \color{violet}\textbf{0.733} \\ 
        06 & \color{blue}\textbf{2.362} & 0.872 & 3.187 & \color{red}\textbf{4.776} & 3.187 & \color{violet}\textbf{1.911} & 1.854 & 1.873 \\ 
        07 & \color{blue}\textbf{24.082} & 0.859 & 5.158 & \color{red}\textbf{6.280} & 5.158 & 2.170 & 2.176 & \color{violet}\textbf{2.302} \\ 
        08 & \color{blue}\textbf{1.926} & 0.384 & 3.504 & 3.493 & \color{red}\textbf{3.504} & 1.15 & 1.141 & \color{violet}\textbf{1.166} \\ 
        09 & \color{blue}\textbf{---} & 2.2 & 1.268 & \color{red}\textbf{5.375} & 1.268 & 1.054 & 1.055 & \color{violet}\textbf{1.086} \\ 
        10 & \color{blue}\textbf{---} & 2.295 & 9.67 & 3.088 & \color{red}\textbf{9.666}& 0.576 & \color{violet}\textbf{0.685} & 0.559 \\ %\hline
    \end{tabular}}
    \caption{Total \gls{aad} $\times 10^{-2}$ across Sequences \texttt{00} to \texttt{10} in SemanticKITTI after masking the nodes from the input graphs (up) and masking the edges at the last layer of SEM-GAT (down). %We follow the same color convention as in \cref{table:attention_accuracy}. 
    The colours indicate the highest discrepancy scores after perturbation. {\color{red}\textbf{Red}} indicates highest scores overall, {\color{blue}\textbf{blue}} highest scores after semantic masking, and {\color{violet}\textbf{purple}} highest scores for each semantic class. For multi-class masking, the symbol "{\color{blue}\textbf{---}}" indicates an error in pose estimation due to insufficient registration points or confidence loss.}
    \label{table:TVE_Adj}
        \vspace{-0.5cm}

\end{table*}

\section{Conclusions}
 % for the challenging task of pose estimation
In this work, we investigated the semantic interpretation of attention scores for identifying key elements in a pointcloud and introduced a methodology to evaluate the fidelity of our findings. Our analysis provides a thorough insight into the validity of using attention as an indicator of semantic importance. Our experimental results indicate a strong correlation between attention weights and model performance, allowing us to draw conclusions on expected model behaviour in diverse environments. In our approach, we identify important semantic features in the environment for graph-attention-based pose estimation models. This methodology can be used to explain the model's performance in correlation with the semantics present. 

% We investigate the semantic interpretation of the attention scores from SEM-GAT. In particular, due to the fact that edge attention weights are calculated based on the node embeddings derived from nearest neighbor aggregations, as described in \cref{subsec:sem_gat}, we investigate whether the morphological features of each class generate more concise and descriptive embeddings which result in them receiving highest importance scores, and thus making the query class more important than others.

%\addtolength{\textheight}{-12cm}   % This command serves to balance the column lengths
                                  % on the last page of the document manually. It shortens
                                  % the textheight of the last page by a suitable amount.
                                  % This command does not take effect until the next page
                                  % so it should come on the page before the last. Make
                                  % sure that you do not shorten the textheight too much.
\bibliographystyle{IEEEtran}
\bibliography{WS2}
\let\cleardoublepage=\clearpage
\end{document}